\documentclass[man,floatsintext,hidelinks]{apa7}

\usepackage[american]{babel}
\usepackage{fvextra}
\usepackage{csquotes}
\usepackage{amsmath}
\usepackage{threeparttable}
\usepackage{graphicx}

\usepackage[T1]{fontenc}

\usepackage[style=apa,sortcites=true,sorting=nyt,natbib=true,backend=biber]{biblatex}
\usepackage{tabularx}
\usepackage{booktabs}
\usepackage{adjustbox}
\usepackage{caption}

\usepackage{float}
\usepackage[utf8]{inputenc}


\title{Quantifying Cultural-Psychological Constructs in Four Centuries of Ottoman Court Registers}

\shorttitle{Historical Psychology in the Ottoman World}

\authorsnames[1, 2, 3]{Sila Mutaf, Enes Yılandiloğlu, Mohammad Atari}

\authorsaffiliations{
    {Department of Psychology, University of Michigan},
    {Department of Digital Humanities, University of Helsinki},
    {Department of Psychological and Brain Sciences, University of Massachusetts Amherst}
}

\authornote{
    Sila Mutaf and Enes Yılandiloğlu contributed equally to this work and share first authorship.\\
    Acknowledgments: We thank Mina Ozkaraduman and Ahmet Ziya Boz for their help with annotating documents. This publication was partly supported through the grant \#63867 from the John Templeton Foundation to Mohammad Atari. \\

    \addORCIDlink{Sila Mutaf}{0009-0000-9213-031X} \\
    \addORCIDlink{Enes Yılandiloğlu}{0009-0009-2764-8011} \\ 
    \addORCIDlink{Mohammad Atari}{0000-0002-4358-7783}
}

\leftheader{Mutaf et al.}

\abstract{Populations vary along cultural dimensions such as individualism, collectivism, and honor. But how can these dimensions be studied in historical societies for which conventional psychological data are unavailable? Here, we introduce a computational text-analytic framework for quantifying psychological constructs in 54,670 Ottoman Turkish court records from 21 Istanbul courts spanning 1513--1911 CE. We develop sentence-embedding models adapted to Ottoman legal language and extend a theory-driven, text-analytic pipeline to capture cultural variation along the dimensions of individualism, collectivism, and honor in this historical corpus. We evaluate our models through multiple methods: document retrieval, semantic textual similarity, questionnaire-item classification, and validation against expert annotations. Our adapted sentence transformer (OTA-BERTurk) achieved a Recall@1 of .843 on the test set for document retrieval, a Mean Reciprocal Rank (MRR) of .806 for semantic textual similarity (meaning the best answer usually appeared as a top result), and classified questionnaire items with 89.58\% accuracy. Document-level performance varied across constructs and methods. Qualitative analysis further showed that high scores reflected historically specific legal and institutional patterns. We publicly release the models, construct representations, scoring pipeline, as well as an interactive web interface to support transparent, reproducible historical-psychological analyses of Ottoman Turkish court records spanning approximately four centuries. These computational tools enable the study of psychological variation across centuries of cultural and institutional change in a diverse empire.
}


\begin{document}

\maketitle


Our psychology leaves a unique trace in our language \citep{pennebaker2003psychological}. When language is preserved in written form, it can thereafter serve as an indirect archive of the minds and cultures that produced it \citep{iliev2015automated}. Because written records emerge from particular cultural and institutional settings, recurring patterns can reveal how societies collectively made sense of the world, categorized people, organized social relationships, and evaluated behavior \citep{boyd2017psychological}. Contemporary computational techniques allow us to excavate these ``cognitive fossils'' from historical texts -- incomplete yet informative proxies for psychology preserved in textual records \citep{atari2023historical,baumard2024cognitive}.

Here, we study three foundational cultural dimensions --- individualism, collectivism, and honor --- in Ottoman court registers. These court records provide an important source for studying social and cultural life in the Ottoman Empire, which spanned more than six centuries (1299--1922 CE) and encompassed three continents. Although historians have studied these materials extensively \citep[e.g.,][]{aykan2020sharia,ekinci2005osmanli}, their potential for large-scale computational analysis is largely untapped \citep{Akca_2025, Cosgel_2025}. The corpus analyzed here comprises 54,670 Latin-script transcribed records drawn from 100 Istanbul qadi registers published through the Istanbul Qadi Registers project \citep{isam2023kadisicilleri}. Individualism and collectivism capture the relative emphasis placed on personal agency, uniqueness, and self-expression, on the one hand, and relational obligation, group harmony, and interdependence, on the other \citep{triandis1988individualism,kitayama2024cultural}. These orientations have been widely used to explain cultural variation in self-construal, emotion, cognition, and social behavior \citep{markus1991culture,oyserman2002rethinking}. Honor constitutes a related but distinct cultural logic organized around personal and family reputation, social respect, and responses to insults or moral transgressions. It is especially relevant to the Turkish cultural context \citep{uskul2019social,uskul2023neither} and has been proposed as an additional dimension of cultural variation that complements individualism and collectivism \citep{oyserman2017culture}. Together, these three constructs provide a theoretical framework for quantifying cultural-psychological expression in Ottoman legal texts.

Importantly, however, these psychological constructs may not transfer unchanged in historical and cultural contexts. Although individualism and collectivism were developed to explain cross-cultural variation, their dominant definitions and measurement instruments emerged largely from contemporary academic psychology in the U.S. \citep{talhelm2026defining} and may carry assumptions about personhood, agency, and group membership that do not map neatly onto Ottoman society \citep[see][]{berry1969crosscultural}. Contemporary formulations of honor likewise cannot be assumed to capture its historical meanings. This problem reflects a broader issue in psychological science \citep{medin2017systems,rad2018toward}, whose empirical and theoretical foundations remain disproportionately based on Western, educated, industrialized, rich, and democratic (WEIRD) populations \citep{henrich2010weird}. Emerging historical approaches in psychology address this issue by treating psychological variation as an outcome of processes unfolding over time and investigating how changing cultural, ecological, and institutional environments shape human psychology \citep{atari2025chronospatial}. Making Ottoman Turkish materials accessible allows researchers to examine the cultural evolution of psychological traits in an understudied, non-WEIRD context and facilitates theoretical refinement.

\subsection{From Historical Corpora to Psychological Measurement}

The expansion of digitized archives has enabled researchers to investigate historical variation in cultural-psychological patterns during periods for which surveys and experiments are unavailable. Early studies used word frequencies in large book collections to trace cultural trends, including changes in collective memory, the pursuit of fame, censorship, and historical epidemiology \citep{michel2011}. This shift from individual texts to aggregate patterns parallels the practice of \textit{distant reading} in literary scholarship, which often uses computational methods to identify recurring patterns and long-term trends across large collections \citep{moretti2013distant}. Word-counting analyses were also used to test specific hypotheses about historical change, such as the vocabulary associated with social obligation, authority, individual choice, and materialism in relation to urbanization \citep{greenfield2013changing}, 
the increasing prominence of individualistic language \citep{twenge2012increases,fan2025name}, changes in emotional expression \citep{acerbi2013expression,masui2025emotions}, and the declining cultural salience of moral and virtue-related language \citep{kesebir2012cultural,wheeler2019twentieth}.

These dictionary-based approaches, however, remain limited to the frequencies of individual terms or predefined groups of words and therefore provide limited information about how meaning varies across contexts and periods \citep{almelhem2026enlightenment}. Diachronic word embeddings addressed this limitation by representing words through their distributional relationships and comparing those representations over time \citep{hamilton2016diachronic}. Historians subsequently evaluated word embeddings as tools for studying conceptual and semantic change \citep{wevers2020digital}. In computational social science, word embedding models have been used to examine the changing cultural meanings of social class \citep{kozlowski2019geometry}, gender and ethnic stereotypes \citep{garg2018word}, and representations of social groups \citep{charlesworth2022historical}. In parallel, methods from Natural Language Processing have been used to quantify psychological constructs, such as personality traits \citep{park2015automatic}, group identity \citep{ashokkumar2022tracking}, depression \citep{eichstaedt2018facebook}, and moral values \citep{kennedy2021moral} from social-media posts. 

Applying contextual embedding methods to historical corpora nevertheless presents methodological challenges. Their reliability depends on the quality of the training corpus, their adaptation to the historical language and genre, and the availability of appropriate validation data \citep{manjavacas2022adapting}. Historical corpora also introduce challenges around transcription errors, uneven corpus composition, and limited annotated data \citep{pettit2016historical,martins2022develop}. These limitations are especially important for low-resource languages. Thus, reliable historical-psychological text analysis requires adaptation to the language and corpus, together with validation against expert judgments or external historical evidence \citep{baumard2024cognitive,feuerriegel2025using,atari2023historical,aiello2019triangulation,kennedy2022text}.

Recent work has taken steps toward addressing some of these challenges by combining psychometric theory with contextual language modeling. \citet{atari2023contextualized} introduced Contextualized Construct Representation (CCR), which uses sentence transformers to represent validated psychometric items (or face-valid descriptions of a psychological construct) and target texts in an embedding space \citep{reimers-gurevych-2019-sentence}. Then, relevance of the construct, or ``loading,'' is estimated by calculating the semantic similarity between the text's and the scale items' vectorized representations. \citet{Chen_2024} adapted CCR to Classical Chinese and developed an indirect supervised contrastive-learning procedure for fine-tuning sentence transformers on the Chinese Historical Psychology Corpus, comprising 21,539 paragraphs from 667 historical works. The authors fine-tuned a Classical Chinese model on two historical corpora. By representing passages holistically, this approach can identify construct-relevant meaning that may not be expressed through individual words. Applying CCR to Ottoman court records, hence, requires adapting sentence representations to a specific legal language, determining how contemporary constructs are represented in that language, and benchmarking the resulting scores against expert judgments. 

\subsection{Ottoman Court Registers as a Psychological Archive}

Ottoman court registers were maintained by a judge (\textit{qad\i}) or their deputy (\textit{naib}) and recorded by court scribes (\textit{k\={a}tib}). They document legal relations involving people from different social positions in the imperial center and provinces and illuminate family, social, economic, administrative, and legal life \citep{ugur2026seriyye,aykan2020sharia}. Their contents include family disputes, marriage, homicide, inheritance divisions, contracts, settlements, guardianship, manumission, taxation, price regulation, guild disputes, endowments, appointments, and administrative orders. 

The registers provide broad access to Ottoman legal and social discourse, but constitute an institutionally mediated rather than fully representative record of Ottoman social life; their contents reflect specific patterns of jurisdiction, court use, and documentary practice. Because scribes generally recorded only case summaries, they provide limited detail about proceedings or the actual enforcement of judgments \citep{ekinci2005osmanli}. Despite these limitations, their breadth has made them important sources for historians, legal scholars, and economists \citep{jennings1978zimmis,cosgel2013selection}. Scholars have edited, transcribed, and digitized court registers from across the Ottoman Empire, including collections from Istanbul \citep{yilmaz2010uskudar}, Diyarbakır \citep{izgoer2013diyarbekir}, Jerusalem \citep{rabaia2015kudus}, and Crimea \citep{yasa2025tracing}. These collections have supported research on Islamic legal institutions, slavery and social status, gender and property, legal agency, and family and household organization \citep{baldwin2017islamic,wagner2024rethinking,peirce2003morality,agmon2004women,doumani2017family}. 

The registers' content makes them especially useful for examining social identity and shared cultural beliefs because Ottoman legal documents frequently identify people through relational markers such as parentage, spouse, neighborhood, occupation, religious community, household, or patronage ties. Their coverage across courts and historical periods creates opportunities for examining variation in legal and cultural-psychological language at a scale that would be difficult to address through close reading alone. 

\subsection{The Present Research}

We develop and evaluate a computational framework for measuring cultural-psychological constructs in Ottoman court records. First, we domain-adapt sentence-transformer and long-context embedding models to Ottoman legal language and evaluate the resulting representations through document retrieval, embedding-space analysis, semantic textual similarity (STS) \citep{cer2017semeval}, and questionnaire-item classification (QIC) \citep{Chen_2024}. Second, we compare three approaches to representing psychological constructs: translated questionnaire items, proxy sentences that elaborate semantically aligned and opposed meanings, and expert-authored definitions. In the court records, constructs such as individualism, collectivism, and honor are rarely named directly; instead, they may be expressed through relationships and actions. Representing each construct through multiple proxy sentences allows the models to identify different semantic expressions even when the records do not contain the modern psychological vocabulary. Third, we rely on these representations to generate document-level scores for individualism, collectivism, and honor, and validate these scores against manual annotations by experts in Ottoman history. Moreover, we compare CCR-based measurement with zero-shot and few-shot classifications produced by open-weight Large Language Models (LLMs) \citep{abdurahman2024perils}. Finally, we qualitatively examine the passages with the highest scores to determine whether the retrieved evidence corresponds to the intended constructs and how those concepts are expressed through legal and institutional terminology. We publicly release the adapted models, construct representations, scoring pipeline, and an open-source web interface for other researchers to test their hypotheses and extend our computational approach.

\section{Method}

\subsection{Corpus and Preprocessing}

\subsubsection{Data source}
The corpus used in this study comprises 100 digitized volumes of Istanbul court registers, known variously as \textit{qadi sicilleri} or \textit{şer‘iyye sicilleri}, with Latin-script transcriptions, put together through a collaboration among the Istanbul Metropolitan Municipality, the Centre for Islamic Studies, and Istanbul Medipol University  \citep{isam2023kadisicilleri}. The whole dataset comprises 54,670 documents totaling 11,497,610 words. The dataset spans from 1513 to 1911 CE, recorded by 21 different courts located in Istanbul. The registers were written in the Perso-Arabic script and drew on Ottoman Turkish, although their linguistic composition varied across courts, periods, and document types \citep{Akca_2025}. The dataset was transcribed to Latin script, digitized, and made publicly accessible by \citet{isam2023kadisicilleri}. We utilize the data provided by \citet{Yuksek_2025}. Each record includes the court title, the year of the issue (both Hijri and Gregorian dates), the title, a summary, the text of the case, and the ID information. The summaries in our data were manually prepared by domain experts during the process of transcription into Latin script. These one-sentence summaries have the names of the main actors and the main act or appeal (see Supplementary Materials for examples of these summaries).  

\subsubsection{Preprocessing and Exclusions}
The original dataset contained 54,670 records. For document-level scoring, we removed 56 records that contained a summary but no main text, leaving 54,614 records. We applied stricter criteria when constructing the model-training subset, excluding 133 records from the original dataset: 116 with fewer than 20 text characters, including the 56 records with no main text; 10 with fewer than 20 characters after markup tags were removed; one with an alphabetic-character ratio below .30; and six with summaries shorter than six characters. These exclusions left 54,537 training-eligible records, of which 52,360 had unique summaries (96.0\%).
The dataset initially contained 24 court-title labels, including alternative names or spellings for the same courts. We consolidated three pairs: \textit{Eyüp Mahkemesi} and \textit{Eyüb Mahkemesi (Havass-ı Refia)}; \textit{Anadolu Sadâreti Mahkemesi} and \textit{Anadolu Sadareti Mahkemesi}; and \textit{Rumeli Sadâreti Mahkemesi} and \textit{Rumeli Sadareti Mahkemesi}. This standardization resulted in 21 unique courts.

\subsubsection{Corpus Temporal Coverage }
\begin{figure}[ht]
    \centering
     \caption{Temporal Coverage and Corpus Size Across Ottoman Courts}
    \label{fig:timeline}
    \includegraphics[width=1\textwidth]{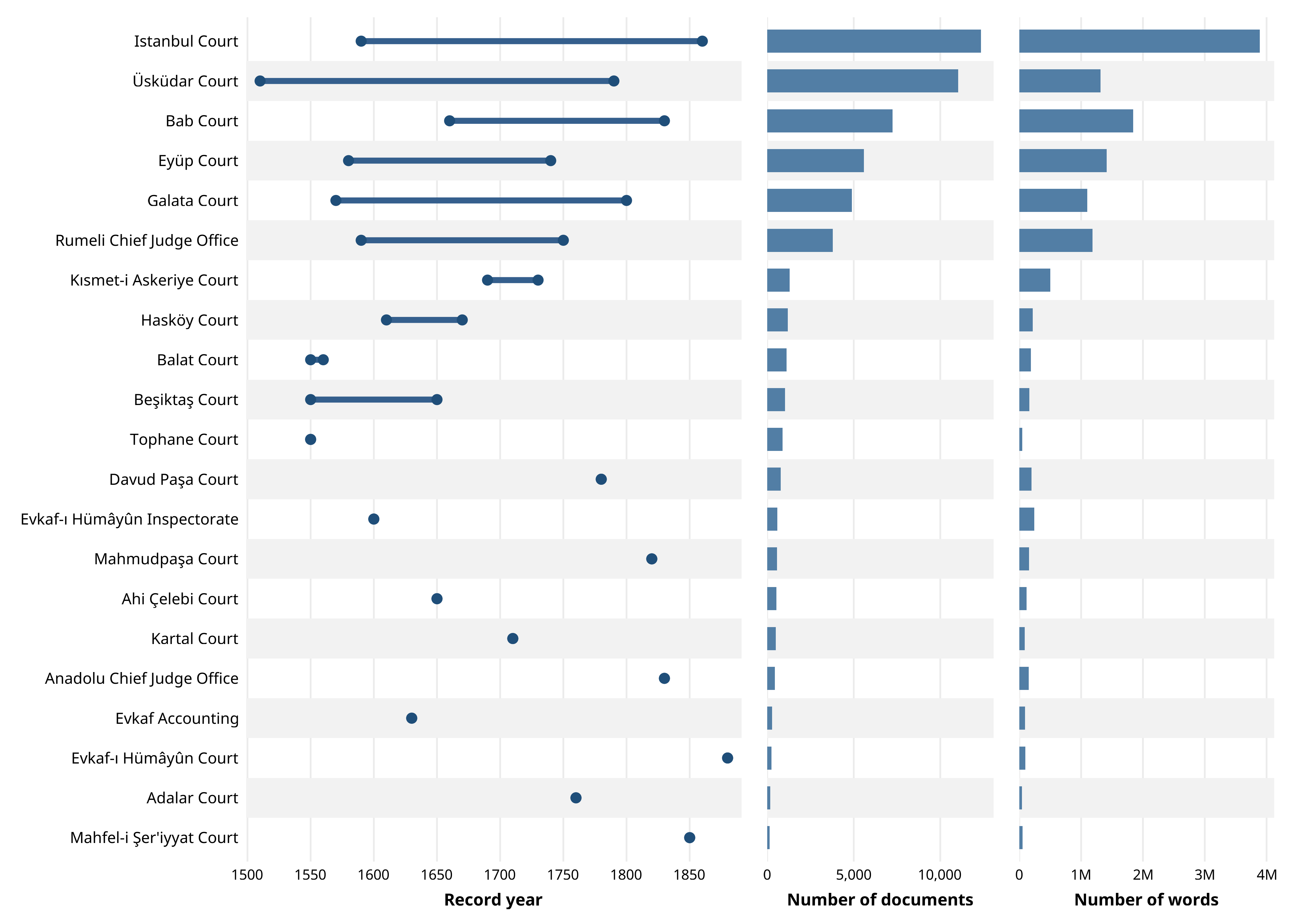}
\end{figure}

The courts are not equally represented over time (see Figure \ref{fig:timeline}). While Bab, Istanbul, Galata, and Üsküdar persisted over decades, other courts appear only for a small period. Across 54,614 documents with text, there are 11,497,610 words, corresponding to an average of 210.5 words per document. The number of documents and words varies notably across courts, as shown in Figure \ref{fig:timeline}.

\subsection{Representing Cultural-Psychological Constructs}
\label{subsec:questionnaires}

We focused on three cultural-psychological constructs: individualism, collectivism, and honor. 

\subsubsection{Questionnaire Items}
Questionnaire items are central to CCR because they provide theoretically grounded semantic anchors for locating a construct within the embedding space \parencite{atari2023contextualized,teitelbaum2025neural}. CCR embeds the questionnaire items and target documents in a shared semantic space and estimates document relevance from their similarity to the resulting construct representation. The content of the questionnaire items therefore determines what the model recognizes as linguistic evidence of a construct. Because validated measures of these constructs are unavailable in Ottoman Turkish, we manually translated each questionnaire item into Latin-transcribed Ottoman Turkish.

\paragraph{Individualism and Collectivism} We represented individualism and collectivism using the 22-item measure developed by \citet{Oyserman_1993}. The measure contains separate sets of items assessing individualistic and collectivistic orientations.

\paragraph{Honor} We represented honor using 25 items drawn from \citet{Mosquera_2002}. The measure includes items assessing moral integrity, family honor, feminine honor, and masculine honor, with one item representing both feminine and family honor.

\subsubsection{Proxy Sentences and Expert Definitions}
 We expanded each item into five aligned and five opposed sentences designed to express semantically aligned and opposed meanings \citep[see][]{teitelbaum2025neural}. We also prepared a single-paragraph expert definition for each construct. Illustrative examples of the questionnaire items, Ottoman Turkish adaptations, aligned and opposed proxy sentences, and expert-authored definitions are provided in Appendix~\ref{app:construct-examples} (Table~\ref{tab:construct-examples}). The complete materials are provided in the Supplementary Materials.

\subsection{Expert-Annotated Evaluation Set}

\subsubsection{Sampling Procedure}

We constructed an expert-annotated evaluation set by sampling 20 documents from each court label represented in the metadata. We sampled records within courts and stratified the selection by decade, excluding documents used in an earlier pilot sample of five records per court. Because two labels for the Eyüp court --- \textit{Eyüp Mahkemesi} and \textit{Eyüb Mahkemesi (Havass-ı Refia)} --- were sampled separately before court names were consolidated, Eyüp contributed 40 documents, whereas each other court contributed 20. Therefore, this procedure resulted in 440 documents.

\subsubsection{Annotation and Label Aggregation}

Three annotators, including a co-first author, independently coded each document for individualism, collectivism, and honor. Annotators followed a detailed codebook containing construct definitions and positive and negative examples drawn from the corpus. They assigned each document a separate label for each construct: present ($1$), absent ($0$), or uncertain ($-1$). All annotators held bachelor's degrees in history or Turkish philology and had relevant knowledge of Ottoman Turkish and training in Ottoman history.
We combined the annotations into a binary document-level label through majority vote. We treated uncertain labels as missing and calculated the majority over the remaining valid labels. For a given construct, we excluded documents when the two remaining valid labels disagreed. This procedure yielded honor labels for 435 documents (31 present and 404 absent), individualism labels for 440 documents (368 present and 72 absent), and collectivism labels for 438 documents (84 present and 354 absent). The complete annotation codebook is provided in the Supplementary Materials.

\subsubsection{Inter-Annotator Agreement}
Based on prior work on psychological annotation \citep{zewail2026camel}, we evaluated inter-annotator agreement using Gwet's AC1, Krippendorff's $\alpha$, Fleiss' $\kappa$, PABAK, and percentage agreement. We treated Gwet's AC1 as the primary coefficient because the prevalence of positive labels differed substantially across constructs, particularly for honor \citep{gwet2008computing,krippendorff2004reliability}.

\subsection{Model Development}

\subsubsection{Data Splits and Evaluation Separation}
After removing two duplicate summary--text pairs, the model-development corpus contained 54,535 records. We held out the 440 expert-annotated documents from all model training and checkpoint selection. We divided the remaining 54,095 records into approximately 80\% training, 10\% validation, and 10\% test sets using a fixed random seed of 42. We performed the split at the summary level, ensuring that records sharing the same summary were assigned to the same subset. We did not stratify the data by court or decade. The resulting subsets contained 43,108 training records, 5,715 validation records, and 5,272 test records. We selected checkpoints using the validation set and conducted document-retrieval and embedding-space evaluations on the held-out test set, using the same records and sampled pairs for every model.

\subsubsection{Fine-tuning Overview}

The training-pair procedure used in earlier historical applications of CCR identifies related documents through the similarity of broad thematic titles \citep{Chen_2024}. The Ottoman court records instead contain one-sentence summaries with personal names and specific legal acts, which can produce high similarity between otherwise unrelated cases. We therefore paired each summary directly with its corresponding court record.

We adapted OTA-XLM-RoBERTa \citep{xlm-roberta-base-ota-kadisicilleri} and BERTurk \citep{schweter_2020} through two training stages. First, we used unsupervised SimCSE \citep{gao-etal-2021-simcse} to obtain sentence-level representations from the masked-language encoders. Second, we trained the models on summary--record pairs using Multiple Negatives Ranking Loss (MNRL), treating other records in the batch as negatives. We adapted the long-context EmbeddingGemma-2 model using the same summary--record objective but without the preliminary SimCSE stage because it was already trained as a sentence- and document-embedding model. The Supplementary Materials provide the complete training procedures, hyperparameters, prompt formats, and implementation settings.

\subsection{Document-Level Construct Scoring}

\subsubsection{Pooling Strategies}

We calculated cosine similarity between each construct representation and the Ottoman court records using three pooling strategies. The whole-document approach encoded the available text of each record as a single vector:

\begin{equation}
\operatorname{Score}_{\mathrm{doc}}(D,q)
=
\cos\left(\mathbf{e}(D),\mathbf{e}(q)\right),
\label{eq:document-score}
\end{equation}

where $D$ denotes a court record and $q$ denotes a questionnaire item, proxy sentence, or definition sentence. Because a construct may be expressed in only part of a record, we also divided each record into sentences. Maximum pooling retained the highest sentence-level similarity:

\begin{equation}
\operatorname{Score}_{\mathrm{max}}(D,q)
=
\max_{s_i \in D}
\cos\left(\mathbf{e}(s_i),\mathbf{e}(q)\right),
\label{eq:max-pooling}
\end{equation}

whereas mean pooling averaged similarity across all sentences:

\begin{equation}
\operatorname{Score}_{\mathrm{mean}}(D,q)
=
\frac{1}{N}
\sum_{i=1}^{N}
\cos\left(\mathbf{e}(s_i),\mathbf{e}(q)\right),
\label{eq:mean-pooling}
\end{equation}

where $N$ is the number of sentences in $D$. Maximum sequence lengths and sentence-segmentation procedures are reported in the Supplementary Materials.

\subsubsection{Construct Representations}

We represented each construct in four ways. \textit{Items} used only the translated questionnaire items. \textit{Aligned} combined each original item with five semantically aligned proxy sentences. \textit{Contrast} additionally incorporated five opposed proxy sentences for each item. \textit{Definition} used the sentences of the expert-authored construct definition.

For a set of construct queries $Q$, we averaged the query-level scores:

\begin{equation}
\operatorname{Score}(D,Q)
=
\frac{1}{|Q|}
\sum_{q \in Q}
\operatorname{Score}(D,q),
\label{eq:query-aggregation}
\end{equation}

where $\operatorname{Score}(D,q)$ was calculated using one of the three pooling strategies described above. For contrast scoring, we subtracted the mean similarity to the opposed query set from the mean similarity to the aligned query set:

\begin{equation}
\operatorname{Score}_{\mathrm{contrast}}(D,C)
=
\operatorname{Score}(D,Q_C^{+})
-
\operatorname{Score}(D,Q_C^{-}),
\label{eq:contrast-score}
\end{equation}

where $Q_C^{+}$ and $Q_C^{-}$ contain the aligned and opposed representations of construct $C$, respectively.

Honor was represented by 25 questionnaire items, whereas individualism and collectivism were each represented by 11 items. Each item was expanded into five aligned and five opposed proxy sentences. The Supplementary Materials provide the original items, Ottoman Turkish adaptations, proxy sentences, and expert-authored definitions.

\subsection{Comparison Methods}
\subsubsection{Embedding Models}
We evaluated the domain-adapted models alongside off-the-shelf masked-language, sentence-embedding, and long-context embedding models. The complete list of models, identifiers, and implementation settings is provided in Appendix~\ref{app:models} (Table~\ref{tab:models}).

\subsubsection{LLM Benchmark}
We evaluated seven open-weight LLMs as automated classifiers of the 440 expert-coded documents. Each model classified the presence ($1$) or absence ($0$) of honor, individualism, and collectivism under both zero-shot and few-shot prompting, resulting in six classifications per document. We used greedy decoding without sampling and disabled model reasoning. We compared the predictions with the expert-derived document labels. The Supplementary Materials provide model identifiers, prompt templates, quantization and inference settings, API-access details, output constraints, and post-processing procedures.

\subsection{Evaluation Tasks}

We evaluated the models at two levels. First, representation-level evaluations assessed whether the embeddings captured general semantic relationships in Ottoman legal language, independently of the expert construct labels. These evaluations included document retrieval, questionnaire-item classification, and semantic textual similarity. Second, document-level validation assessed whether the embedding-based CCR scores and LLM classifications corresponded to expert annotations of honor, individualism, and collectivism.

\subsubsection{Document Retrieval}
We evaluated summary--record alignment on the held-out test split. Each case summary served as a query, and the model ranked the 5,272 test records according to cosine similarity. A record was considered relevant when it corresponded to the query summary. When multiple records shared the same summary, all corresponding records were treated as relevant. We used the same queries, candidate records, and relevance judgments for every model. We report Recall@1, Recall@5, Recall@10, MRR@10, and NDCG@10. Recall@\(k\) measures the proportion of all records associated with a query summary that appear among the first \(k\) retrieved records, averaged across queries. MRR@10 reflects the rank of the first relevant record, whereas NDCG@10 evaluates the ranking of all relevant records within the first ten positions.

\subsubsection{Questionnaire-Item Classification}
The QIC task evaluates how effectively sentence embeddings capture the boundaries between psychological concepts. Using an SVM classifier on item-level vector representations of the questionnaire, this evaluation demonstrates the capability of domain-adapted sentence transformer models to reliably distinguish psychological constructs. More specifically, we analyze how well a given model classifies the items in the questionnaires. We used each questionnaire item and its five aligned proxy sentences as separate samples (282 samples from 47 items: 150 honor, 66 individualism, 66 collectivism). Items were classified with a support vector machine (RBF kernel, $C = 1.0$, $\gamma = \text{scale}$) with balanced class weights, using fixed hyperparameters without additional tuning. We used 10-fold cross-validation grouped by questionnaire item (GroupKFold), thus an item and its proxy sentences never appear in both the training and the test folds. Reported values are the means across the 10 folds, and the values after $\pm$ are the standard deviations across folds. All items in the honor scale from \citet{Mosquera_2002} are labeled as `honor,' while items from \citet{Oyserman_1993} are labeled as either `individualism' or `collectivism.'
For the EmbeddingGemma models, the items and proxy sentences were encoded with the \texttt{Document} prompt.

\subsubsection{Semantic Textual Similarity}

In addition to QIC, we implemented a pairwise STS evaluation task to assess how the models distinguish semantically related from unrelated sentence pairs in Ottoman Turkish legal texts. First, we prepared 100 semantically similar within-topic pairs in Ottoman Turkish on legal topics such as debt, land dispute, and embezzlement. Ottoman Turkish sentences cover a wide range of styles, from heavy legal language to plain prose that stays close to modern Turkish. As none of these sentences were in the training data of the models, this dataset serves as novel, unseen data for evaluation. The data can be accessed through Supplementary Materials.

Then we constructed 19,900 pairwise combinations from the 200 sentences, consisting of 100 semantically similar within-topic pairs (positive class) and 19,800 cross-topic dissimilar pairs (negative class). Cosine similarity scores were extracted for each pair. We report two kinds of metrics. First, retrieval metrics: each sentence is used as a query against its matched sentence, and vice versa, and we report the Mean Reciprocal Rank (MRR) and Recall@1 with 95\% bootstrap confidence intervals over topics. Second, ROC-AUC: the cross-register AUC uses the 100 positives and only the negatives, and the all-pairs AUC uses all 19{,}900 pairs. We also report the area under the precision--recall curve (PR-AUC), because 100 positives are outnumbered by 19{,}800 negatives, and ROC-AUC can look high even when precision at a usable threshold is low. For the EmbeddingGemma models, both sides of a pair (and of a questionnaire-item pair in QIC) were encoded with the \texttt{Document} prompt, because both sides are the same kind of object. Other prompts are reported in the Supplementary Materials. We compare models with paired bootstrap differences over topics.

\subsubsection{Comparative Document-Level Validation}

We evaluated two approaches to document-level construct measurement against the expert-derived labels. The embedding-based approach produced continuous CCR scores for honor, individualism, and collectivism. The zero-shot and few-shot LLM approaches produced binary predictions indicating whether each construct was present ($1$) or absent ($0$). All methods were evaluated on the same expert-annotated documents, which were excluded from model training and checkpoint selection.

For the continuous embedding scores, we report ROC-AUC with 95\% confidence intervals obtained through 1,000-iteration bootstrap resampling. We additionally report average precision because the prevalence of positive labels differed substantially across constructs: 7\% for honor, 84\% for individualism, and 19\% for collectivism. Pearson's $r$ and Spearman's $\rho$ are reported in the Supplementary Materials.

For the binary LLM predictions, we report balanced accuracy, F1, precision, sensitivity, and specificity. For a hard binary predictor, ROC-AUC is numerically equivalent to balanced accuracy. Nevertheless, continuous embedding scores provide more detailed ranking information than binary LLM outputs, so direct differences between the two approaches should be interpreted cautiously. We conducted all analyses separately for honor, individualism, and collectivism.

\subsection{Qualitative Examination of High-Scoring Records}

To interpret what the document-level construct scores captured, we ranked the records separately by their individualism, collectivism, and honor scores, calculated using OTA-BERTurk with contrast scoring and whole-document embeddings, and qualitatively examined the top-10 scoring records per construct. For each record, we considered the document type, the actors and relationships represented, and the passages most closely associated with the construct. The upper tail of the score distribution differed substantially across constructs. For example, only 35 individualism records but 208 honor records scored more than three standard deviations above their respective corpus means. We therefore used rank rather than a standardized-score threshold to define the primary qualitative sample. For honor, we additionally included records that are outside the top-10 sample that nevertheless received extreme scores (i.e., higher than three SDs above the mean). 

\subsection{Transparency and Openness}

We report how the corpus and expert-annotated evaluation sample were constructed, all data exclusions, all construct representations, and all evaluation procedures used in this study. The original Latin-script transcriptions of the Ottoman court records are accessible through the Istanbul Qadi Registers project at \url{https://kadisicilleri.org}. Due to copyright restrictions, we do not redistribute the complete record texts.
The domain-adapted models are available at https://zenodo.org/records/24282664. The analysis and model-training code, scoring pipeline, preprocessing procedures, and code used to reproduce the tables and figures are available at https://zenodo.org/records/24282664. The translated questionnaire items, aligned and opposed proxy sentences, expert-authored construct definitions, annotation codebook, LLM prompts, expert annotations, and evaluation materials are available in the Supplementary Materials and at https://zenodo.org/records/24282664. The interactive web interface is available at https://koras0ff.github.io/ottoman-court-explorer/.
Of note, neither the study design nor the analyses were preregistered.

\subsection{Use of Generative Artificial Intelligence in Manuscript Preparation}

During manuscript preparation, the authors used ChatGPT \citep{openai2026chatgpt} and Claude \citep{anthropic2026claude} to assist with language editing and \LaTeX{} formatting. The authors reviewed and revised all AI-assisted material and independently verified the accuracy of factual claims and citations. This use was limited to manuscript preparation and was distinct from the open-weight LLM benchmark described above. The tool was not used to generate or modify the study data or expert annotations, conduct statistical analyses, or determine the reported results.

\section{Results}

\subsection{Model Performance Evaluation}

\subsubsection{Document Retrieval}

Table~\ref{tab:retrieval} presents the document-retrieval results on the held-out model test set. The SimCSE stage alone did not substantially improve retrieval: NDCG@10 remained .005 for the XLM-RoBERTa model and .024 for the BERTurk model. Performance improved substantially after summary--record training, with NDCG@10 reaching .889 for OTA-XLM-RoBERTa, .898 for OTA-BERTurk, and .955 for OTA-EmbeddingGemma-2. Among the domain-adapted models, OTA-EmbeddingGemma-2 achieved the strongest overall document-retrieval performance.

\begin{table}[!t]
\centering
\caption{Document Retrieval Performance on the Test Set}
\label{tab:retrieval}

\small
\renewcommand{\arraystretch}{1.12}

\begin{tabular*}{\textwidth}{
    @{\extracolsep{\fill}}
    llccccc
    @{}
}
\toprule
\textbf{Model}
& \textbf{Stage}
& \textbf{NDCG@10}
& \textbf{MRR@10}
& \textbf{R@1}
& \textbf{R@5}
& \textbf{R@10} \\
\midrule

OTA XLM-RoBERTa
& Base
& 0.003 & 0.002 & 0.002 & 0.003 & 0.004 \\

BERTurk
& Base
& 0.011 & 0.009 & 0.006 & 0.013 & 0.019 \\

EmbeddingGemma-2
& Base
& 0.566 & 0.528 & 0.449 & 0.627 & 0.692 \\

\addlinespace
SimCSE (XLM-RoBERTa)
& Stage 1
& 0.005 & 0.003 & 0.001 & 0.005 & 0.012 \\

SimCSE (BERTurk)
& Stage 1
& 0.024 & 0.018 & 0.011 & 0.029 & 0.041 \\

\addlinespace
OTA-XLM-RoBERTa
& Stage 2
& 0.889 & 0.873 & 0.831 & 0.920 & 0.939 \\

OTA-BERTurk
& Stage 2
& 0.898 & 0.883 & 0.843 & 0.925 & 0.946 \\

OTA-EmbeddingGemma-2
& Stage 2
& \textbf{0.955}
& \textbf{0.947}
& \textbf{0.918}
& \textbf{0.974}
& \textbf{0.984} \\

\bottomrule
\end{tabular*}

\vspace{4pt}

\begin{minipage}{\textwidth}
\footnotesize
\textit{Note.}
Bold values indicate the highest performance in each column. NDCG@10 = normalized discounted cumulative gain at 10; MRR@10 = mean reciprocal rank at 10; R@1, R@5, and R@10 = recall at ranks 1, 5, and 10, respectively.
\end{minipage}

\end{table}

\subsubsection{Questionnaire-Item Classification}

We evaluated model performance across a 3-construct classification scheme (individualism, collectivism, and honor). Table \ref{tab:qic_3constructs} summarizes accuracy, macro F1, and weighted F1 for the masked language models and our domain-adapted models. The performance of off-the-shelf embedding models is reported in the Supplementary Materials. Among our models, OTA-BERTurk performed best ($89.58\%$ accuracy), outperforming its base model BERTurk ($85.08\%$). Several larger general-purpose embedding models were higher, such as Octen-Embed 8B ($93.50\%$). OTA-XLM-RoBERTa ($80.17\%$) was far above the XLM-RoBERTa masked language models ($47.17\%$), which score at about the share of the largest class (honor, 53\%). Fine-tuning did not improve EmbeddingGemma-2: OTA-EmbeddingGemma-2 scored $85.25\%$ with the \texttt{Document} prompt, against $87.00\%$ for the base model, and it performed worse under every prompt (Supplementary Materials). A plausible reason is that the summary--text objective teaches the model to match short summaries to long records, which is a different task than classifying short questionnaire items.

\begin{table}[!t]
\centering
\caption{Model performance on the QIC task.}
\label{tab:qic_3constructs}
\small
\begin{tabularx}{\textwidth}{>{\raggedright\arraybackslash}X c c c}
\toprule
\textbf{Model} & \textbf{Accuracy (\%)} & \textbf{F1-Macro (\%)} & \textbf{F1-Weighted (\%)} \\
\midrule
\multicolumn{4}{l}{\textit{\textbf{Fill-Mask / MLM Encoders}}} \\
XLM-RoBERTa-base & 47.17 $\pm$ 14.42 & 24.49 $\pm$ 5.86 & 33.21 $\pm$ 12.42 \\
OTA XLM-RoBERTa & 47.17 $\pm$ 13.80 & 24.38 $\pm$ 4.47 & 33.12 $\pm$ 12.20 \\
BERTurk & 85.08 $\pm$ 7.74 & 77.74 $\pm$ 12.37 & 83.96 $\pm$ 8.51 \\
SerBERT & \textbf{85.25 $\pm$ 8.35} & \textbf{78.10 $\pm$ 12.74} & \textbf{84.06 $\pm$ 9.20} \\
\addlinespace
\multicolumn{4}{l}{\textit{\textbf{Domain-adapted Models (ours)}}} \\
OTA-XLM-RoBERTa & 80.17 $\pm$ 7.72 & 72.09 $\pm$ 12.92 & 78.99 $\pm$ 8.65 \\
OTA-BERTurk & \textbf{89.58 $\pm$ 5.57} & \textbf{83.81 $\pm$ 9.70} & \textbf{88.72 $\pm$ 6.24} \\
OTA-EmbeddingGemma-2 & 85.25 $\pm$ 7.42 & 79.15 $\pm$ 10.52 & 84.52 $\pm$ 7.79 \\
\bottomrule
\end{tabularx}

\vspace{6pt}

\begin{minipage}{\textwidth}
\footnotesize
\textit{Note.}
Bold values indicate the best result within each model family. Accuracy is the percentage of samples assigned to the correct class. Macro F1 is the unweighted mean of the class-specific F1 scores. Weighted F1 averages the class-specific F1 scores according to class size. 
\end{minipage}
\end{table}

\begin{table}[!htbp]
\centering
\caption{Performance on the Semantic Textual Similarity Task}
\label{tab:sts_performance}

\footnotesize
\setlength{\tabcolsep}{4pt}

\resizebox{\textwidth}{!}{%
\begin{tabular}{@{}lccccc@{}}
\toprule
\textbf{Model}
& \textbf{MRR [95\% CI]}
& \textbf{R@1}
& \textbf{Cross-reg. AUC [95\% CI]}
& \textbf{AUC (all)}
& \textbf{PR-AUC} \\
\midrule

\multicolumn{6}{l}{\textit{\textbf{MLM encoders}}} \\
XLM-RoBERTa-base
& 0.139 [0.105, 0.177]
& 0.045
& 0.623 [0.603, 0.648]
& 0.486
& 0.005 \\

OTA XLM-RoBERTa
& 0.351 [0.290, 0.408]
& 0.240
& 0.771 [0.733, 0.809]
& 0.679
& 0.016 \\

BERTurk
& 0.331 [0.276, 0.389]
& 0.235
& 0.683 [0.657, 0.713]
& 0.590
& 0.009 \\

SerBERT
& \textbf{0.566 [0.502, 0.632]}
& \textbf{0.465}
& \textbf{0.870 [0.841, 0.900]}
& \textbf{0.820}
& \textbf{0.075} \\

\addlinespace
\multicolumn{6}{l}{\textit{\textbf{Domain-adapted models}}} \\
OTA-XLM-RoBERTa
& 0.762 [0.696, 0.823]
& 0.665
& \textbf{0.977 [0.965, 0.987]}
& \textbf{0.973}
& 0.419 \\

OTA-BERTurk
& \textbf{0.806 [0.740, 0.866]}
& \textbf{0.745}
& 0.972 [0.956, 0.985]
& 0.967
& \textbf{0.521} \\

OTA-EmbeddingGemma-2
& 0.762 [0.692, 0.832]
& 0.690
& 0.961 [0.936, 0.980]
& 0.953
& 0.334 \\

\bottomrule
\end{tabular}%
}

\vspace{4pt}

\begin{minipage}{\textwidth}
\footnotesize
\textit{Note.}
Bold values indicate the best result
within each model family. MRR = mean reciprocal rank; R@1 = recall at 1; Cross-reg. AUC = cross-register area under the receiver operating characteristic curve; AUC (all) = area under the receiver operating characteristic curve calculated across all pairs; PR-AUC = area under the precision--recall curve.
\end{minipage}

\end{table}

\subsubsection{Semantic Textual Similarity}

Performance on the semantic textual similarity task varied considerably across model families (Table~\ref{tab:sts_performance}). The two largest general-purpose embedding models were nearly perfect: Qwen3-Embed 8B (MRR = .968, cross-register AUC = .999) and Octen-Embed 8B (MRR = .965, cross-register AUC = .998). As their intervals overlap, we do not rank them. Among our domain-adapted models, OTA-BERTurk was the strongest (MRR = .806 [.740, .866]), followed by OTA-EmbeddingGemma-2 (MRR = .762) and OTA-XLM-RoBERTa (MRR = .762). Paired bootstrap differences (MRR) among the three adapted models were not significant (OTA-BERTurk versus OTA-EmbeddingGemma-2: $\Delta$MRR $= .043$ $[-.022, .104]$; versus OTA-XLM-RoBERTa: $.044$ $[-.006, .095]$). Thus, STS scores do not clearly separate them alone. Domain adaptation clearly helped each base model as EmbeddingGemma-2 went from an MRR of .607 to .762 (cross-register AUC from .897 to .961), and the OTA-XLM-RoBERTa masked language model went from .351 to .762. The choice of prompt made no significant difference for OTA-EmbeddingGemma-2 (Supplementary Materials). Precision stays modest. Yet, the PR-AUC is .521 for OTA-BERTurk and .334 for OTA-EmbeddingGemma-2, even though their ROC-AUCs are above .90. In Table \ref{tab:sts_performance}, the best value within each model category is marked in bold; the off-the-shelf embedding models are reported in the Supplementary Materials.

\subsection{Expert-Annotation Reliability}
Agreement varied considerably across the three constructs (Table~\ref{tab:agreement_scores}). For honor, annotations reached 86\% observed agreement, and Gwet's AC1 was 0.89, indicating substantial agreement. The low Krippendorff $\alpha$ (0.37) and Fleiss $\kappa$ (0.40) for honor can be explained by the skew in the annotations rather than genuine disagreement. Individualism is the most reliable construct, with high agreement across every metric (AC1 = 0.89, $\alpha$ = 0.71, $\kappa$ = 0.71, PABAK = 0.83; 88\% raw agreement). Agreement estimates for collectivism varied across metrics (AC1 = 0.53; $\alpha$ = 0.17; $\kappa$ = 0.17), with raw agreement of 55\%. This divergence may partly reflect substantial differences in annotators' use of the positive label: annotator 1 assigned it to 205 of 440 items, compared with 29 and 73 items for annotators 2 and 3, respectively.

\begin{table}[!htbp]
\centering
\begin{threeparttable}

\caption{Inter-Annotator Agreement Scores}
\label{tab:agreement_scores}

\small

\begin{tabular*}{\textwidth}
{@{\extracolsep{\fill}}lccccc@{}}
\toprule
\textbf{Construct} &
\textbf{Gwet's AC1} &
\textbf{Krippendorff's $\alpha$} &
\textbf{Fleiss' $\kappa$} &
\textbf{PABAK} &
\textbf{\% Agreement} \\
\midrule

Honor
& 0.89 [0.86, 0.92]
& 0.37 [0.27, 0.47]
& 0.40
& 0.81
& 86\% \\

Individualism
& 0.89 [0.85, 0.91]
& 0.71 [0.63, 0.78]
& 0.71
& 0.83
& 88\% \\

Collectivism
& 0.53 [0.46, 0.59]
& 0.17 [0.11, 0.23]
& 0.17
& 0.40
& 55\% \\

\bottomrule
\end{tabular*}

\begin{tablenotes}[flushleft]
\footnotesize
\item \textit{Note.}
AC1 = Gwet's first-order agreement coefficient;
$\alpha$ = Krippendorff's alpha;
$\kappa$ = Fleiss' kappa;
PABAK = prevalence-adjusted bias-adjusted kappa.
Values in brackets are 95\% confidence intervals.
\% Agreement indicates percentage agreement among the annotators.
\end{tablenotes}

\end{threeparttable}
\end{table}

\subsection{Document-Level Validation Against Expert Annotations}
\label{sec:PM}

Table \ref{tab:doc_level} shows the ROC-AUC of the MLM encoders and our domain-adapted models under one fixed configuration (the contrast score with whole-document embeddings) and the balanced accuracy of the four LLMs discussed below under zero-shot and few-shot prompting; the off-the-shelf embedding models and the remaining LLMs are in the Supplementary Materials. No single approach consistently outperformed the others across all constructs. 

\begin{table}[!htbp]
\centering
\begin{threeparttable}

\caption{Document-Level Validation Against Expert Annotations}
\label{tab:doc_level}

\footnotesize
\setlength{\tabcolsep}{4pt}

\begin{tabular*}{\textwidth}{
@{\extracolsep{\fill}}
lccc
@{}
}
\toprule
\textbf{Model} &
\textbf{Honor} &
\textbf{Individualism} &
\textbf{Collectivism} \\
\midrule

\multicolumn{4}{l}{\textit{\textbf{MLM encoders (ROC-AUC)}}} \\

XLM-RoBERTa-base
& 0.572 [0.451, 0.682]
& 0.519 [0.427, 0.606]
& 0.350 [0.288, 0.417] \\

OTA XLM-RoBERTa
& \textbf{0.697 [0.599, 0.790]}
& 0.280 [0.212, 0.356]
& 0.606 [0.542, 0.672] \\

BERTurk
& 0.349 [0.252, 0.460]
& \textbf{0.712 [0.639, 0.783]}
& \textbf{0.627 [0.558, 0.694]} \\

SerBERT
& 0.424 [0.312, 0.536]
& 0.705 [0.634, 0.778]
& 0.500 [0.419, 0.578] \\

\addlinespace

\multicolumn{4}{l}{\textit{\textbf{Domain-adapted models (ours; ROC-AUC)}}} \\

OTA-XLM-RoBERTa
& 0.687 [0.585, 0.787]
& 0.424 [0.340, 0.509]
& \textbf{0.727 [0.663, 0.793]} \\

OTA-BERTurk
& 0.654 [0.563, 0.741]
& 0.541 [0.463, 0.616]
& 0.664 [0.589, 0.737] \\

OTA-EmbeddingGemma-2
& \textbf{0.697 [0.587, 0.796]}
& \textbf{0.556 [0.475, 0.637]}
& 0.529 [0.452, 0.611] \\

\addlinespace

\multicolumn{4}{l}{
\textit{\textbf{Open-weight LLMs (balanced accuracy; zero-shot / few-shot)}}
} \\

DeepSeek-V4-Flash
& \textbf{0.677} / 0.510
& \textbf{0.560} / 0.682
& 0.673 / \textbf{0.653} \\

Qwen 3.5 9B
& 0.532 / 0.537
& 0.519 / 0.778
& 0.560 / 0.526 \\

Qwen 3.8 27B
& 0.669 / 0.602
& 0.512 / \textbf{0.833}
& \textbf{0.740} / 0.632 \\

Gemma 4-E4B-it
& 0.615 / \textbf{0.605}
& 0.522 / 0.768
& 0.652 / 0.595 \\

\bottomrule
\end{tabular*}

\begin{tablenotes}[flushleft]
\footnotesize
\item \textit{Note.}
Values for MLM encoders and domain-adapted embedding models are ROC-AUCs with 95\% bootstrap confidence intervals. Results for off-the-shelf embedding models and the remaining LLMs are reported in the Supplementary Materials.
\end{tablenotes}

\end{threeparttable}
\end{table}

\subsubsection{Individualism}

Detecting individualism was challenging for our domain-adapted embedding models, whereas few-shot LLMs performed better. Among the MLM encoders, BERTurk achieved the highest ROC-AUC (.712 [.639, .783]), while Jina Embeddings v5 Small was the strongest off-the-shelf embedding model (.710 [.636, .781]; see Supplementary Materials). Among our models, OTA-EmbeddingGemma-2 performed best, although its ROC-AUC remained close to chance (.556 [.475, .637]); OTA-XLM-RoBERTa performed below chance (.424). LLM performances varied and notably depended on the prompting strategy. Zero-shot balanced accuracy ranged from .496 to .560, whereas few-shot Qwen 3.8 27B reached .833, followed by Qwen 3.5 9B (.778) and Gemma 4-E4B-it (.768). 

\subsubsection{Collectivism}
For collectivism, our models fared well. OTA-XLM-RoBERTa had the highest AUC of all embedding models (.727 [.663, .793]), although its interval overlaps with that of the best off-the-shelf model, Octen-Embed 8B (.666 [.600, .730]; Supplementary Materials). OTA-BERTurk (.664) was on par with it, but OTA-EmbeddingGemma-2 stayed close to chance (.529). BERTurk was the best MLM encoder (.627 [.558, .694]). The strongest LLM was zero-shot Qwen 3.8 27B (balanced accuracy .740), slightly above our best AUC, although the two metrics cannot be compared directly. Collectivism also has the least reliable labels of the three constructs, so these scores should be interpreted with caution, and tested in the future with larger labeled data.

\subsubsection{Honor}
Honor is the construct where the embedding models did best. The best off-the-shelf model, Qwen3-Embed 8B, scored .800 [.708, .890] (Supplementary Materials). The MLM encoders were behind: OTA XLM-RoBERTa was the best of them (.697 [.599, .790]). Our three domain-adapted models have similar scores, OTA-EmbeddingGemma-2 on top (.697 [.587, .796]), then OTA-XLM-RoBERTa (.687) and OTA-BERTurk (.654). That puts the best off-the-shelf model about .10 above our best one. The intervals overlap, though, and with only 31 positive documents we cannot separate the groups with high confidence. The LLMs did worse than the embedding models on honor: zero-shot DeepSeek-V4-Flash reached a balanced accuracy of .677, and zero-shot Qwen 3.8 27B was slightly less accurate (.669).

\subsection{Qualitative Examination of High-Scoring Records}

\subsubsection{Individualism}
The 10 highest-scoring individualism records were all related to manumission (see Table~\ref{tab:qual-individualism}). These documents identify enslaved persons through their names, origins, physical characteristics, and legal status before formally recording their transition to freedom. In one case, Mehmed Çelebi declares that he has freed Ruhısâni, an enslaved woman of Russian origin, for the sake of God and divine approval; the court subsequently recognizes her as free, like other freeborn adult women \parencite[judgment no.~573, fol.~85b, entry~3, p.~428]
{yilmaz2010galata20}. Another record documents, through an authorized representative and witnesses, the manumission of Şaban, an enslaved man of European origin \parencite[judgment no.~32, original-text no.~17-3, p.~51]
{yilmaz2011galata5}. Importantly, each deed formally reclassifies a named person from property to a free legal subject. The model therefore appears to associate individualism primarily with legal individuation through emancipation, which is the formal recognition of a person as possessing a legal status separate from that of a former owner. 

\subsubsection{Collectivism}

The highest-scoring collectivism records were dominated by religious endowment deeds (see Table~\ref{tab:qual-collectivism}). In one of the highest-scoring examples, Musa Efendi dedicates a residence first to Kadızâde Şeyh Mehmed Efendi, then to his own descendants, and ultimately to a member of the Naqshbandi order or a poor Muslim \parencite[judgment no.~94, fol.~17a, entry~1, p.~120]
{yilmaz2019rumeli40}. Successive occupants must administer, maintain, and repair the property. When Musa Efendi attempts to revoke the endowment, the trustee refuses and the court declares it binding. The model thus appears to associate collectivism with the transformation of individual wealth into a collective institution that links family, religious community, and future beneficiaries through shared responsibilities and intergenerational obligations.

\subsubsection{Honor}
The highest-scoring honor records were dominated by manumission deeds and other disputes over freedom and legal status (see Table~\ref{tab:qual-honor}). In one case, Fatıma Hanım formally frees Sertâbe, an enslaved woman of Russian origin, and the court records that she has joined the ranks of freeborn women and that her former owner retains no claim over her beyond the legally recognized bond of patronage \parencite[judgment no.~160, fols.~33a--33b, entry~4, p.~176]{yilmaz2019bab11}. Other records that are high-scoring (i.e., more than 3 SDs above the mean)  link honor more directly to the public evaluation of conduct. A record, which was not on the top 10-highest scoring cases, but was still 3.47 SDs above the mean, describes a group identified by the court as Roma who admitted drinking wine and were accused of insulting janissaries and directing improper remarks toward Muslim families. Community members further portrayed their behavior as habitual and socially disruptive, and these allegations were entered into the register \parencite[judgment no.~372, fol.~35a, entry~4, p.~250]{yilmaz2010uskudar84}. 
The model therefore appears to associate honor both with the formal recognition of social and legal standing and with communal judgments of morally appropriate conduct. This interpretation is consistent with accounts of Ottoman honor as a relational process shaped through publicly negotiated reputation \citep{peirce2014honor}. However, this interpretation warrants caution because the highest-scoring records were dominated by manumission deeds. The pattern may reflect the formulaic language of legal status, freedom, and pious motivation more than the broader historical connotations of honor.

\subsection{Interactive Corpus Exploration Using the Web Interface}
\begin{figure}[!htbp]
    \centering
    \caption{An Example of Cultural-Psychological Construct Loadings, Displayed on the Web Interface}
    \label{fig:interface_scores}
    \includegraphics[width=\linewidth]{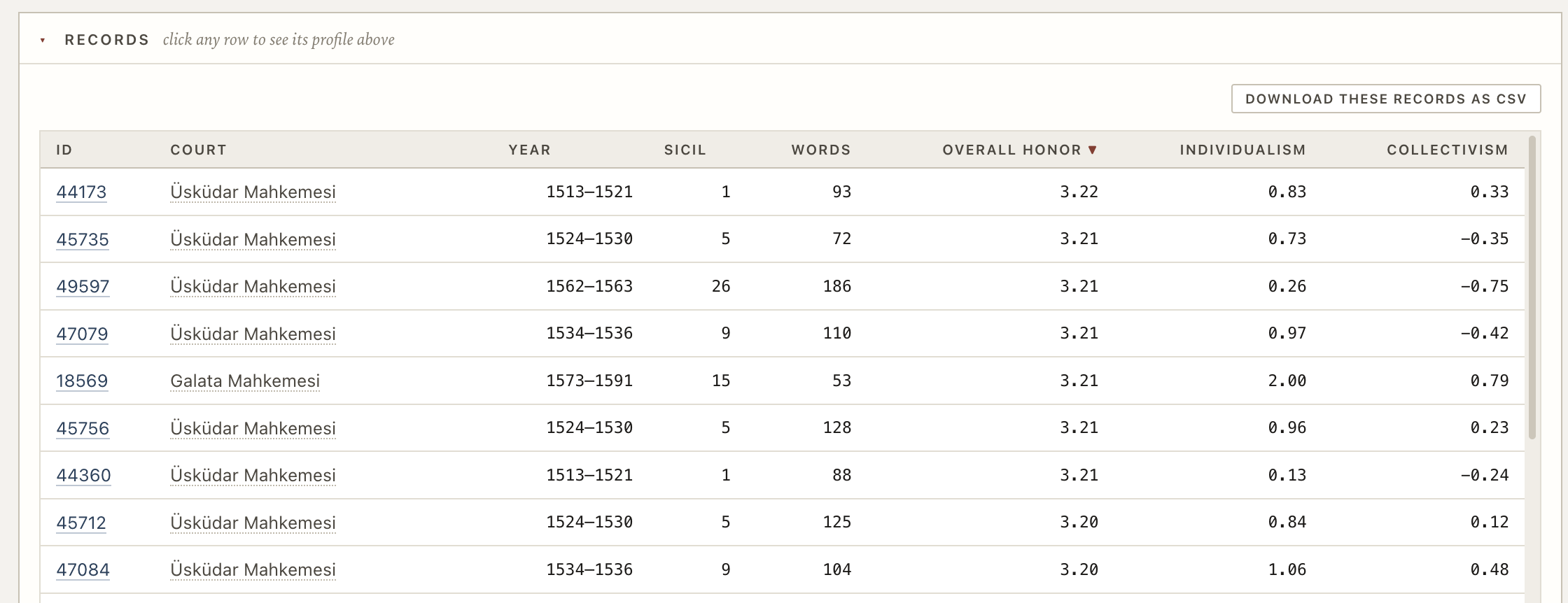}
\end{figure}

As a publicly available research resource, we release an open-access web interface to analyze 54,670 Ottoman court records to quantify other psychological constructs. The outputs in the interface are calculated using the OTA-BERTurk model with contrastive scoring across the full document, the same approach used to generate results in Table \ref{tab:doc_level}. The interface provides the outputs yielded by the CCR method (an example shown in Figure \ref{fig:interface_scores}). The interface aggregates vector similarity scores for each psychological construct, providing a visualization of semantic trends across decades and court districts. Users can view the outputs generated by both the base and fine-tuned embedding models. Additionally, users can select which constructs they want to view in the visualizations tab. 

\begin{figure}[!htbp]
    \centering
    \caption{An Example of Psychological Variation Over Time, Displayed on the Web Interface.}
    \label{fig:change}
    \includegraphics[width=\linewidth]{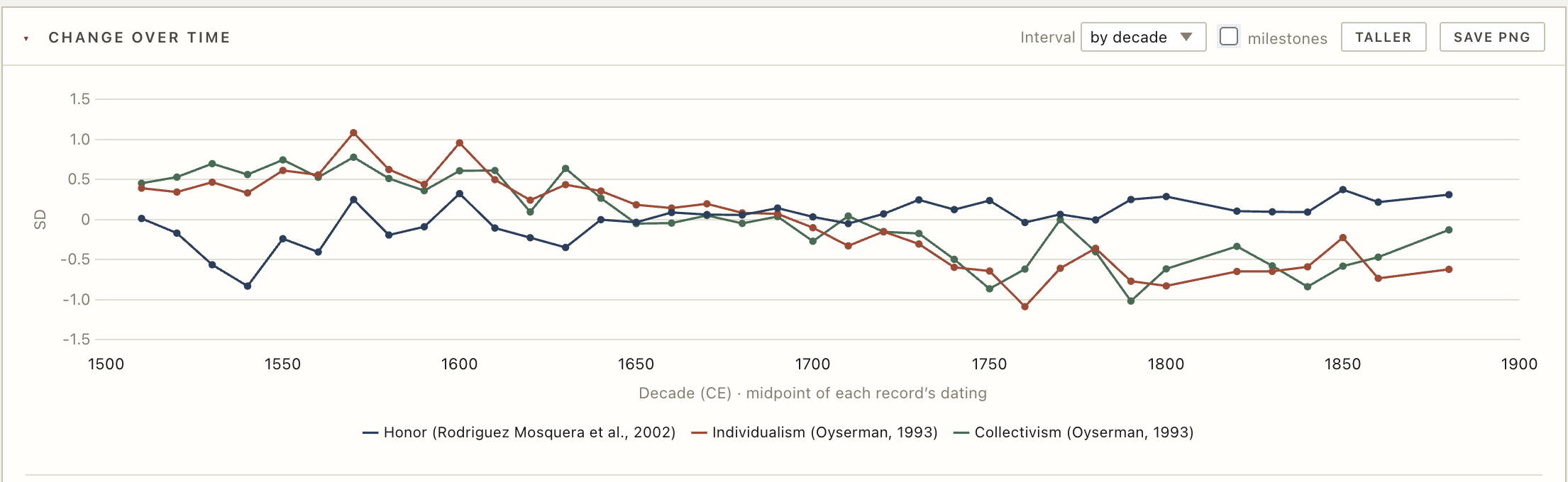}
\end{figure}

The interface presents both high-level statistical visualizations and court records themselves. This allows researchers to inspect standardized construct scores over time. Users can compare constructs across decades to identify historical variation (see Figure \ref{fig:change}).  
We use embedding-based scores in the interface rather than LLM annotation because the interface covers all 54,670 records. An LLM would need six prompts per record (three constructs times two prompt types) over more than 50,000 records, which is slow and expensive, and raises reproducibility concerns \citep{mcloughlin2026proprietary}. Embedding the records once and scoring them by cosine similarity is fast and reproducible, since readers can inspect which sentences drive a score up or down.

\section{Discussion}

We develop a domain-adapted computational framework for analyzing Ottoman court records and demonstrate its application with examples to historical-psychological measurement. Following the idea that written records can serve as indirect archives of the psychological orientations of the populations that produced them \citep{atari2023historical}, we treat the Ottoman court registers as a site where individual and collective identity can be examined empirically. The framework identifies and compares the linguistic and institutional forms through which cultural-psychological constructs, in this case individualism, collectivism, and honor, were expressed in the legal discourse of the historical Ottoman state. 

Across 54,670 records from 100 court-register volumes, we adapted sentence-embedding models to Ottoman legal language and evaluated the resulting representations across several tasks. Domain adaptation generally improved summary--record alignment, document retrieval, and semantic textual similarity, although the levels of improvement varied across models and tasks, and larger off-the-shelf models sometimes performed better. Document-level performance depended jointly on the construct, model family, construct representation, and pooling strategy: no single model or scoring approach performed best across the constructs of interest. 

The results for individualism and collectivism revealed different strengths and limitations. For individualism, the domain-adapted embedding models performed weakly against the expert labels. For collectivism, OTA-XLM-RoBERTa achieved the strongest performance among the embedding models. The qualitative results for high-scoring individualism and collectivism documents were conceptually coherent. High-scoring individualism records were consistently manumission records that show legal individuation, whereas high-scoring collectivism records centered on endowments that dedicated private property to religious and communal purposes, creating lasting systems of shared responsibility and intergenerational obligation. This systematic retrieval suggests that the pipeline identified historically situated textual patterns aligned with the intended constructs, even though quantitative performance varied across models and operationalizations. Close reading of high-scoring records further shows why historical interpretation must accompany computational scoring.

Honor produced a comparatively complex pattern across the quantitative and qualitative evaluations. At the document level, the embedding models distinguished honor more successfully than the other constructs: Qwen3-Embed 8B achieved a ROC-AUC of .800, while the strongest domain-adapted model, OTA-EmbeddingGemma-2, reached .697. The open-weight LLMs performed less strongly, although their balanced accuracies are not directly comparable with the ROC-AUCs calculated from continuous embedding scores. These results indicate that the construct representations captured a signal corresponding to the expert honor labels. Close reading, however, showed that the highest-scoring records were dominated by manumission deeds, suggesting that the models may have responded strongly to their language of publicly recognized status, freedom, pious motivation, and moral self-presentation. Records more directly concerning reputation and communal evaluations of conduct also received high scores but appeared less frequently at the extreme upper end of the distribution. 

Future research should evaluate honor within documentary genres and develop construct representations grounded more directly in historically situated concepts. We used the Honor Scale developed by \citet{Mosquera_2002}, a well-established cross-cultural measure that has also been adapted and validated in Turkish samples \citep{ceylanbatur2019honor}. Nevertheless, honor is not a unitary concept, and the scale's dimensions may not cover all the meanings expressed in Ottoman legal discourse. In the Ottoman state, honor was commonly employed by terms such as \textit{ırz} and \textit{namus} and operated through relational processes of public reputation even when it was not named directly \citep{peirce2014honor}. Contemporary related terms do not necessarily denote the same phenomenon: research on Turkish usage distinguishes among \textit{namus}, \textit{şeref}, and \textit{onur}, associating them with different aspects of sexual and familial reputation, public standing, and dignity \citep{dilmac2014our}. These meanings also changed historically; for example, \textit{ırz} acquired specifically gendered and legal meanings through nineteenth-century Ottoman penal codification \citep{tugonaran2022codification}. Future work could therefore construct separate representations of these concepts that are period-sensitive. 

This result supports treating contemporary constructs as analytical starting points rather than historically invariant categories. Repeated retrieval of a narrower, historically coherent pattern may reveal legal or institutional configuration that aligns with one component of the contemporary construct, or how part of that contemporary construct was organized differently in the given discourse. At the same time, interpretive coherence alone does not establish validity. The proposed proxy-sentence method is most useful when it remains traceable, meaning researchers can inspect the sentences that define a construct, examine the passages that drive a document's score, compare alternative scales or expert definitions, and revise the representation when it retrieves an unintended legal or linguistic pattern. This iterative movement is precisely what the framework is designed to support. 

\subsection{Limitations}

Several limitations should be acknowledged. First, the registers are institutionally mediated documents written by court scribes, not transparent reports of private beliefs, everyday speech, or population-level prevalence. Their contents reflect who used the courts, which disputes entered their jurisdiction, how cases were summarized, and which documentary formulas scribes employed. Construct scores should consequently be interpreted as properties of recorded legal discourse, not direct estimates of the psychological traits of Ottoman populations. 

Second, the expert benchmark is small relative to the corpus, and agreement varies markedly by construct. High agreement for individualism supports firmer conclusions, whereas weaker agreement for collectivism makes model-human correspondence more difficult to interpret. Honor presents the additional problem of strong prevalence imbalance, which can yield high raw agreement and Gwet's AC1 alongside much lower kappa-type coefficients. Honor has only 31 positive documents, so its intervals are wide, and the cross-document analysis (131 positive pairs) and the STS set (100 pairs) are small. Our comparison between the sentence transformers (256 tokens) and EmbeddingGemma-2 (2,048 tokens) also mixes model and context length. Finally, pooling choices can reverse a construct's ranking, thus document-level scores depend on the configuration.

Third, as we have discussed before, the constructs examined here are not culturally or historically neutral. Although individualism and collectivism were developed to explain cross-cultural variation, their dominant definitions and measurement instruments emerged largely from contemporary academic psychology and may carry assumptions that do not transfer to Ottoman society \citep{berry1969crosscultural}. As discussed, honor may have greater contextual relevance given its importance in Turkish cultural psychology, but contemporary formulations of honor likewise cannot be assumed to capture its historical meanings. Moreover, the questionnaire items, proxy sentences, and expert definitions used to represent these constructs shape what the models can recognize as relevant evidence. Construct scores should therefore be treated as properties of the specific discourse rather than direct measurements of psychological traits. Therefore close-reading should complement quantitative scores when possible. 

\subsection{Future Directions}

The study of historical texts opens up new possible questions to be asked and answered by social science researchers, and especially for cultural psychologists who are interested in cultural change. Previous research has used historical literature to trace changes in romantic love \citep{baumard2022cultural}, prosociality \citep{martins2020rise}, contemporary psychological variation and historical information about Church exposure and kinship institutions to explain contemporary psychological variation in individualism, conformity and impersonal prosociality \citep{schulz2019church}. Similarly, our framework opens new avenues for examining how cultural-psychological patterns varied across time and geography -- in this case, across four centuries and regions on three continents over which the Ottoman Empire exercised authority. Temporal analyses can investigate whether such cultural-psychological constructs became more or less prominent during periods such as imperial expansion or political decline, and look at how specific administrative reforms, economic crises, or certain demographic changes have affected the cultural-psychological constructs reflected in historical writings. Comparisons with contemporaneous archives from other societies could then examine whether such patterns were specific to the Ottoman context or reflected broader developments occurring across states during the same historical period. Such analyses can also generate historically grounded hypotheses about the origins of contemporary cultural-psychological patterns, which can then be tested using data from living populations. 

Our framework also supports questions about demographic diversity. The 21 courts represented in the corpus include courts serving Muslim-majority and non-Muslim-majority neighborhoods (see Supplementary Materials, Table S1, for the complete list), as well as courts with specialized institutional jurisdictions. Across these different social settings, the courts formed part of a shared judicial apparatus through which the Ottoman state exerted its authority. Researchers can therefore ask whether courts used different linguistic patterns when recording the claims of different communities or represented agency and moral responsibility differently across social settings. Such comparisons may reveal the diverse moral worlds and forms of social life that operated in the historical neighborhoods of the Ottoman Empire.

Much can be done to strengthen and refine the present framework and improve its results. Future work should expand and adjudicate the expert-coded sample, report construct-specific error analyses, validate scores against independent historical classifications, and test robustness across construct definitions and pooling strategies. Diachronic and cross-court analyses should use balanced samples or multilevel models and should be paired with close readings of the records that drive apparent differences. The web interface can support this work by allowing researchers to move from aggregate patterns to the underlying passages. Used in this way, computational scores can accompany historical interpretation by providing a reproducible method for locating, comparing, and refining hypotheses about culturally and historically situated forms of cultural-psychological constructs.

\section{Conclusion}

Ultimately, this application to Ottoman court registers broadens the borders of both psychology and history. We introduced a reproducible computational framework that combines the construction of a 54,670 record corpus, the domain adaptation of language models to Ottoman legal discourse which was evaluated through semantic textual similarity and questionnaire-item classification, and document-level validation against expert annotations. To represent psychological constructs, we compared different opperationalizations of the same construct, namely questionnaire items, proxy sentences, and expert-authored definitions. We complemented these quantitative evaluations with close readings of high-scoring records to identify the textual and institutional patterns underlying the model outputs. By making the corpus, adapted models, construct representations, scoring pipeline, and web interface publicly available, we provide a reusable methodological infrastructure for historical-psychological research in Ottoman Empire. 
The framework extends psychological inquiry beyond living respondents and modern survey instruments while complementing close historical reading with corpus-scale analysis. In doing so, it expands the temporal, cultural, and methodological scope of social-scientific research and creates opportunities to investigate how culturally meaningful representations of agency, obligation, status, and reputation varied across courts and historical periods.

\printbibliography

\appendix

\section{Models Evaluated}
\label{app:models}

\begin{table}[htbp]
\centering
\caption{Embedding Models Evaluated}
\label{tab:models}

\footnotesize
\setlength{\tabcolsep}{3pt}
\renewcommand{\arraystretch}{1}

\begin{tabularx}{\textwidth}{
@{}
>{\raggedright\arraybackslash}p{0.22\textwidth}
>{\raggedright\arraybackslash}X
>{\centering\arraybackslash}p{0.08\textwidth}
>{\centering\arraybackslash}p{0.09\textwidth}
>{\raggedright\arraybackslash}p{0.16\textwidth}
@{}
}
\toprule
\textbf{Model} &
\textbf{Identifier or base model} &
\textbf{Params} &
\shortstack{\textbf{Max.}\\\textbf{length}} &
\textbf{Adaptation} \\
\midrule

\multicolumn{5}{l}{\textit{MLM encoders}} \\
XLM-RoBERTa-base &
\nolinkurl{FacebookAI/xlm-roberta-base} &
278M & 256 & None \\
OTA XLM-RoBERTa &
\nolinkurl{enesyila/ota-roberta-base-kadisicilleri} &
278M & 256 & MLM on corpus \\
BERTurk &
\nolinkurl{dbmdz/bert-base-turkish-cased} &
111M & 256 & None \\
SerBERT &
\nolinkurl{sumeyyeakca/serbert-ottoman-turkish-bert-model} &
111M & 256 & None \\
\addlinespace
\multicolumn{5}{l}{\textit{Off-the-shelf embedding models}} \\
EmbeddingGemma-2 &
\nolinkurl{google/embeddinggemma-2} &
740M & 2,048 & None \\
Jina Embeddings v5 Nano &
\nolinkurl{jinaai/jina-embeddings-v5-text-nano} &
212M & 1,024 & None \\
Jina Embeddings v5 Small &
\nolinkurl{jinaai/jina-embeddings-v5-text-small} &
596M & 1,024 & None \\
Qwen3-Embed 4B &
\nolinkurl{Qwen/Qwen3-Embedding-4B} &
4.02B & 8,192 & None \\
Qwen3-Embed 8B &
\nolinkurl{Qwen/Qwen3-Embedding-8B} &
7.57B & 8,192 & None \\
Octen-Embed 4B &
\nolinkurl{Octen/Octen-Embedding-4B} &
4.02B & 8,192 & None \\
Octen-Embed 8B &
\nolinkurl{Octen/Octen-Embedding-8B} &
7.57B & 8,192 & None \\
\addlinespace
\multicolumn{5}{l}{\textit{Domain-adapted}} \\
OTA-XLM-RoBERTa &
OTA XLM-RoBERTa &
278M & 256 & SimCSE + summary--text MNRL \\
OTA-BERTurk &
BERTurk &
111M & 256 & SimCSE + summary--text MNRL \\
OTA-EmbeddingGemma-2 &
EmbeddingGemma-2 &
740M & 2,048 & Summary--text MNRL \\

\bottomrule
\end{tabularx}
\end{table}

\section{Examples of Construct Representations}
\label{app:construct-examples}

\begin{table}[htbp]
\centering
\caption{Illustrative Examples of Construct Representations}
\label{tab:construct-examples}

\footnotesize
\setlength{\tabcolsep}{4pt}
\renewcommand{\arraystretch}{0.75}

\begin{tabularx}{\textwidth}{
@{}
>{\raggedright\arraybackslash}p{0.15\textwidth}
>{\raggedright\arraybackslash}p{0.20\textwidth}
>{\raggedright\arraybackslash}X
@{}
}
\toprule
\textbf{Construct} &
\textbf{Representation} &
\textbf{Illustrative text} \\
\midrule

\textit{Honor}
& Original item
& "... your family had a bad reputation?" \\

& Ottoman Turkish adaptation
& "... aileniz kötü bir şöhrete sahip olsaydı?" \\

& Aligned proxy
& Hânedanımızın nâm-ı nişanına halel gelmesi, cümlemiz için ölümden beter bir züldür. \\

& Opposed proxy
& Ailenin mazideki hataları, ferdin istikbaline mani teşkil etmemelidir. \\

& Expert-definition excerpt
& Şeref; şahsiyet, etvâr, hissiyât ve saâdet-i şahsiyyedeki ihtilâfâtı îzâh içün istimâl olunan dîğer bir temel mefhûmdur. \\

\addlinespace[4pt]

\textit{Individualism}
& Original item
& "I am unique, different from everyone else." \\

& Ottoman Turkish adaptation
& "Ben ferîdim, herkesten farklıyım." \\

& Aligned proxy
& "Cümleden müstesnâ bir şahsiyetim vardır." \\

& Opposed proxy
& "Kimseden bir imtiyâzım yoktur." \\

& Expert-definition excerpt
& Ferdiyetperver harslarda, nefs üzerine daha ziyâde teberrüz vardır ve eşhâs, mensûp oldukları büyük zümrenin menâfiinden ziyâde kendi menâfi-i şahsiyyelerini mukaddem tutarlar. \\

\addlinespace[4pt]

\textit{Collectivism}
& Original item
& "To know who I really am, you must see me with members of my group." \\

& Ottoman Turkish adaptation
& "Mensûb olduğum zümreleri bilirseniz, hakîkatte kim olduğumu da bilirsiniz." \\

& Aligned proxy
& "Zümrem bilinirse asıl mâhiyetim de idrak olunur." \\

& Opposed proxy
& "Dâhil olduğum zümreler şahsiyetimin özünü tarif etmez." \\

& Expert-definition excerpt
& Cemiyetperver harslardaki eşhâs, zümreleri tarafından tahmîl edilen kavâid ve rollere imtisâl etmeğe meyyâl olup, kendi ihtiyâçlarından ziyâde başkalarının ihtiyâçlarına teveccüh ederler. \\

\bottomrule
\end{tabularx}

\par\vspace{3pt}
\begin{minipage}{\textwidth}
\footnotesize
\raggedright
\textit{Note.} The examples illustrate the structure of the construct
representations. Document-level scoring used the complete sets of questionnaire
items, aligned proxies, opposed proxies, and expert-definition sentences.
Complete materials are provided in the Supplementary Materials.
\end{minipage}

\end{table}

\section{Records Discussed in the Qualitative Examination}
\label{app:qualitative-records}

\begin{table}[htbp]
\centering
\caption{Highest-Scoring Individualism Records Examined Qualitatively}
\label{tab:qual-individualism}

\footnotesize
\setlength{\tabcolsep}{4pt}
\renewcommand{\arraystretch}{1}

\begin{tabularx}{\textwidth}{
@{}
>{\centering\arraybackslash}p{0.06\textwidth}
>{\centering\arraybackslash}p{0.09\textwidth}
>{\raggedright\arraybackslash}p{0.22\textwidth}
>{\centering\arraybackslash}p{0.13\textwidth}
>{\raggedright\arraybackslash}X
@{}
}
\toprule
\textbf{Rank} &
\shortstack{\textbf{SD above}\\\textbf{mean}} &
\textbf{Court} &
\textbf{Date} &
\textbf{Source locator} \\
\midrule

1 & 3.56 & Galata Court & 1575--1576 &
Register 5; vol. 32, p. 51; judgment 32; original text 17-3 \\

2 & 3.48 & Istanbul Court & 1661--1663 &
Register 10; vol. 51, p. 438; judgment 473; original text 59a-2 \\

3 & 3.48 & Balat Court & 1563 &
Register 2; vol. 11, p. 263; judgment 412; original text 73a-2 \\

4 & 3.44 & Galata Court & 1596--1599 &
Register 20; vol. 35, p. 393; judgment 521; original text 78a-2 \\

5 & 3.41 & Istanbul Court & 1591--1617 &
Register 191; vol. 44, p. 180; judgment 167; original text 34b-4 \\

6 & 3.40 & Rumeli Chief Judge Office Court & 1633 &
Register 56; vol. 14, p. 63; judgment 24; original text 7b-1 \\

7 & 3.38 & Galata Court & 1573--1591 &
Register 15; vol. 34, p. 72; judgment 60; original text 11b-4 \\

8 & 3.30 & Balat Court & 1557--1558 &
Register 1; vol. 41, p. 139; judgment 186; original text 28b-2 \\

9 & 3.30 & Galata Court & 1575--1576 &
Register 5; vol. 32, p. 51; judgment 34; original text 18-1 \\

10 & 3.30 & Galata Court & 1596--1599 &
Register 20; vol. 35, p. 428; judgment 573; original text 85b-3 \\

\bottomrule
\end{tabularx}

\par\vspace{3pt}
\begin{minipage}{\textwidth}
\footnotesize
\raggedright
\textit{Note.} SD above mean indicates the number of standard deviations by
which a record's individualism score exceeded the corpus-wide mean. All 10
records concern manumission. The published texts are subject to copyright restrictions.
\end{minipage}

\end{table}

\begin{table}[p]
\centering
\caption{Highest-Scoring Collectivism Records Examined Qualitatively}
\label{tab:qual-collectivism}

\footnotesize
\setlength{\tabcolsep}{4pt}
\renewcommand{\arraystretch}{1}

\begin{tabularx}{\textwidth}{
@{}
>{\centering\arraybackslash}p{0.06\textwidth}
>{\centering\arraybackslash}p{0.09\textwidth}
>{\raggedright\arraybackslash}p{0.23\textwidth}
>{\centering\arraybackslash}p{0.13\textwidth}
>{\raggedright\arraybackslash}X
@{}
}
\toprule
\textbf{Rank} &
\shortstack{\textbf{SD above}\\\textbf{mean}} &
\textbf{Court} &
\textbf{Date} &
\textbf{Source locator} \\
\midrule

1 & 4.98 &
Rumeli Chief Judge Office Court &
1656--1658 &
Register 106; vol. 50, p. 599; judgment 626; original text 95b-1 \\

2 & 4.97 &
Rumeli Chief Judge Office Court &
1623--1624 &
Register 40; vol. 46, p. 120; judgment 94; original text 17a-1 \\

3 & 4.76 &
Evkaf-ı Hümâyûn Inspectorate Court &
1608--1626 &
Register 1; vol. 45, p. 358; judgment 311; original text 96a-1 \\

4 & 4.63 &
Eyüp Court &
1585--1587 &
Register 3; vol. 22, p. 181; judgment 282; original text 30b-1 (Arabic) \\

5 & 4.58 &
Rumeli Chief Judge Office Court &
1633 &
Register 56; vol. 14, p. 39; judgment 1; original text 1b-1 \\

6 & 4.58 &
Eyüp Court &
1644 &
Register 49; vol. 26, p. 114; judgment 103; original text 19a-1 \\

7 & 4.54 &
Istanbul Court &
1661--1663 &
Register 10; vol. 51, p. 256; judgment 220; original text 27a-1 \\

8 & 4.54 &
Rumeli Chief Judge Office Court &
1656--1658 &
Register 106; vol. 50, p. 533; judgment 560; original text 81b-2 \\

9 & 4.54 &
Evkaf-ı Hümâyûn Inspectorate Court &
1608--1626 &
Register 1; vol. 45, p. 506; judgment 449; original text 145b-1 \\

10 & 4.53 &
Istanbul Court &
1663--1664 &
Register 12; vol. 16, p. 341; judgment 379; original text 33b-1 \\

\bottomrule
\end{tabularx}

\par\vspace{3pt}
\begin{minipage}{\textwidth}
\footnotesize
\raggedright
\textit{Note.} SD above mean indicates the number of standard deviations by
which a record's collectivism score exceeded the corpus-wide mean. All 10
records concern religious endowments, including endowments of money, land,
residences, and other property. The published texts are subject to copyright restrictions.
\end{minipage}

\end{table}

\begin{table}[p]
\centering
\caption{Highest-Scoring Honor Records Examined Qualitatively}
\label{tab:qual-honor}

\footnotesize
\setlength{\tabcolsep}{4pt}
\renewcommand{\arraystretch}{1}

\begin{tabularx}{\textwidth}{
@{}
>{\centering\arraybackslash}p{0.06\textwidth}
>{\centering\arraybackslash}p{0.09\textwidth}
>{\raggedright\arraybackslash}p{0.22\textwidth}
>{\centering\arraybackslash}p{0.13\textwidth}
>{\raggedright\arraybackslash}X
@{}
}
\toprule
\textbf{Rank} &
\shortstack{\textbf{SD above}\\\textbf{mean}} &
\textbf{Court} &
\textbf{Date} &
\textbf{Source locator} \\
\midrule

1 & 4.13 &
Bab Court &
1670--1671 &
Register 11; vol. 53, p. 354; judgment 425; original text 80a-2 \\

2 & 4.08 &
Bab Court &
1670--1671 &
Register 11; vol. 53, p. 176; judgment 160; original text 33a-4 \\

3 & 4.08 &
Bab Court &
1670--1671 &
Register 11; vol. 53, p. 176; judgment 158; original text 33a-2 \\

4 & 4.01 &
Rumeli Chief Judge Office Court &
1656--1658 &
Register 106; vol. 50, p. 505; judgment 524; original text 76a-5 \\

5 & 4.00 &
Rumeli Chief Judge Office Court &
1704 &
Register 161; vol. 59, p. 472; judgment 491; original text 119b-1 \\

6 & 3.84 &
Galata Court &
1596--1599 &
Register 20; vol. 35, p. 396; judgment 527; original text 78b-3 (Arabic) \\

7 & 3.83 &
Galata Court &
1606--1607 &
Register 32; vol. 36, p. 59; judgment 47; original text 9a-3 (Arabic) \\

8 & 3.81 &
Rumeli Chief Judge Office Court &
1633 &
Register 56; vol. 14, p. 200; judgment 204; original text 39a-2 (Arabic) \\

9 & 3.79 &
Bab Court &
1666--1667 &
Register 3; vol. 17, p. 781; judgment 1020; original text 121a-2 \\

10 & 3.78 &
Istanbul Court &
1801--1803 &
Register 78; vol. 82, p. 234; judgment 251; original text 33b-3 \\

\addlinespace[4pt]
\multicolumn{5}{l}{\textit{Additional high-scoring cases}} \\

34 & 3.47 &
Galata Court &
1663 &
Register 90; vol. 40, p. 192; judgment 216; original text 28b-2 \\

34 & 3.47 &
Üsküdar Court &
1590--1591 &
Register 84; vol. 10, p. 250; judgment 372; original text 35a-4 \\

70 & 3.31 &
Üsküdar Court &
1737--1738 &
Register 396; vol. 68, p. 249; judgment 279; original text 74a-1 \\

\bottomrule
\end{tabularx}

\par\vspace{3pt}
\begin{minipage}{\textwidth}
\footnotesize
\raggedright
\textit{Note.} SD above mean indicates the number of standard deviations by
which a record's honor score exceeded the corpus-wide mean. Records are ordered
from highest to lowest honor score. The published texts are subject to copyright restrictions.
\end{minipage}

\end{table}

\end{document}